\documentclass[runningheads]{llncs}

\usepackage[T1]{fontenc}
\usepackage{graphicx}
\usepackage{booktabs}
\usepackage{array}
\usepackage{amsmath,amssymb}
\usepackage{xcolor}
\usepackage[capitalize]{cleveref}
\usepackage{url}
\begin{document}

\title{SAGE: Semantic Explainability of\\ Attention-Based Survival Models in Computational Pathology}
\titlerunning{Semantic Explainability of Survival Models}

\author{Abdallah Lamane\inst{1,2} \and Abdul Rahman Diab\inst{2} \and Ren-Chin Wu\inst{3} 
        \and William Lotter\inst{2,4,5}}
\authorrunning{A. Lamane et al.}
\institute{Massachusetts Institute of Technology, Cambridge, MA, USA \and  
    Department of Data Science, Dana-Farber Cancer Institute, Boston, MA \and 
    Department of Pathology, Dana-Farber Cancer Institute, Boston, MA \and
    Department of Pathology, Brigham and Women’s Hospital, Boston, MA \and
    Harvard Medical School, Boston, MA
\\
    \email{lotterb@ds.dfci.harvard.edu}}

\maketitle

\begin{abstract}
Attention-based multiple instance learning (ABMIL) is the predominant approach for slide-level prediction in computational pathology, yet its attention maps provide only local explanations: they indicate where a model focuses but not which histological features drive its predictions or how the model behaves across a patient cohort. We present Semantic Attention Global Explanations (SAGE), a post-hoc framework that extracts global, language-grounded explanations from a frozen ABMIL model. Using a pathology vision-language model, SAGE scores image patches against a dictionary of 25 histological concepts, aggregates these scores according to the model's learned attention, and quantifies how each concept relates to prediction risk across a cohort. Applied to survival prediction using seven TCGA cancer cohorts and three foundation models, SAGE recovered established prognostic features, such as the adverse association of necrosis, while revealing cancer-specific biology, including a favorable angiogenic signature in renal cell carcinoma consistent with known molecular subtypes. Ablation studies demonstrated that these associations depend on the model's learned attention rather than concept prevalence alone, and that the concept dictionary captures much of the prognostic information encoded by the foundation model features. Through semantically-grounded explanations, SAGE provides a scalable, model-agnostic framework for understanding what ABMIL survival models learn, enabling pathologists to interpret model behavior at the cohort level and offering the potential for biomarker identification. 

\end{abstract}

\section{Introduction}

Attention-based multiple instance learning (ABMIL)~\cite{ilse2018attention} is a standard approach for weakly supervised tasks in computational pathology, including emerging applications in survival prediction directly from H\&E whole-slide images (WSIs). As these models move toward clinical decision support,  explainability is essential for adoption. ABMIL offers \emph{local, spatial} explainability through attention
heatmaps for individual slides, but these maps highlight only regions of interest
and leave the underlying \emph{biological semantics}, namely the histological concepts the model
associates with favorable or adverse outcomes, opaque.

In addition to semantic explainability, there is a need for \emph{global} explanations that
go beyond individual slides to characterize what a model has learned overall. Such global views can help
check biological grounding, facilitate trust, and
support knowledge discovery. Concept-bottleneck and concept-guided approaches \cite{Koh2020-ti,sun2025conceptmil,liu2025vlsa}
address this by training models to predict through predefined human-interpretable
concepts, but they typically require extensive concept annotations and have limited
flexibility once trained.

We present a global semantic explainability framework for ABMIL
survival models that does not require retraining or additional annotations. Our
approach leverages pathology-specific vision-language models (VLMs) to map ABMIL
attention weights onto a dictionary of pathology concepts, and then aggregates
concept scores across slides into a cohort-level explainability profile. We apply
our framework to seven TCGA cohorts and three pathology foundation models, where we find that the recovered semantic profiles capture aspects of known biology, with both pan-cancer and cancer-specific prognostic trends. 

\vspace{-3pt}
\section{Related Work}
\vspace{-3pt}
\paragraph{Pathology foundation models.}
CONCH~\cite{lu2024conch} and MUSK~\cite{xiang2025musk} are pathology VLMs trained
to align histology images
with text captions in a shared embedding space. Vision-only foundation models, such as UNI2~\cite{chen2024uni}, are commonly trained using self-distillation and contrastive self-supervised learning objectives.

\paragraph{Concept- and language-grounded explainability in MIL.}
A growing line of work incorporates textual concepts into the WSI
classifier and trains it end-to-end. ConceptMIL~\cite{sun2025conceptmil} is a
label-free concept bottleneck model whose prediction is a linear combination of
VLM-predicted concepts on the top attended patches; ConcepPath~\cite{zhao2024conceppath}
extracts expert concepts with GPT-4 and learns a concept-guided hierarchical
attention; VLEER~\cite{nguyen2025vleer} concatenates retrieved text embeddings
to the visual features before training an MIL aggregator. All three focus on classification tasks rather than survival, and their concept vocabularies are directly incorporated into the predictive representation or model training pipeline.

\paragraph{Vision-language survival analysis.}
VLSA~\cite{liu2025vlsa} performs survival prediction
with language: it encodes ordinal prognostic text priors and uses them as
auxiliary signals that guide instance aggregation during training. Conversely, our framework is designed to explain existing, frozen survival models, making it applicable to vision-only
attention-MIL predictors not designed with text in mind. Moreover, rather than imposing
language priors and subsequently determining their attribution scores, SAGE reads out concept associations from a model's attention, allowing different vocabularies to be queried retrospectively
using the same model. 
\vspace{-4pt}
\section{Semantic Attention Global Explanations (SAGE)}
\vspace{-4pt}
\begin{figure}[t]
\centering
\includegraphics[width=\textwidth]{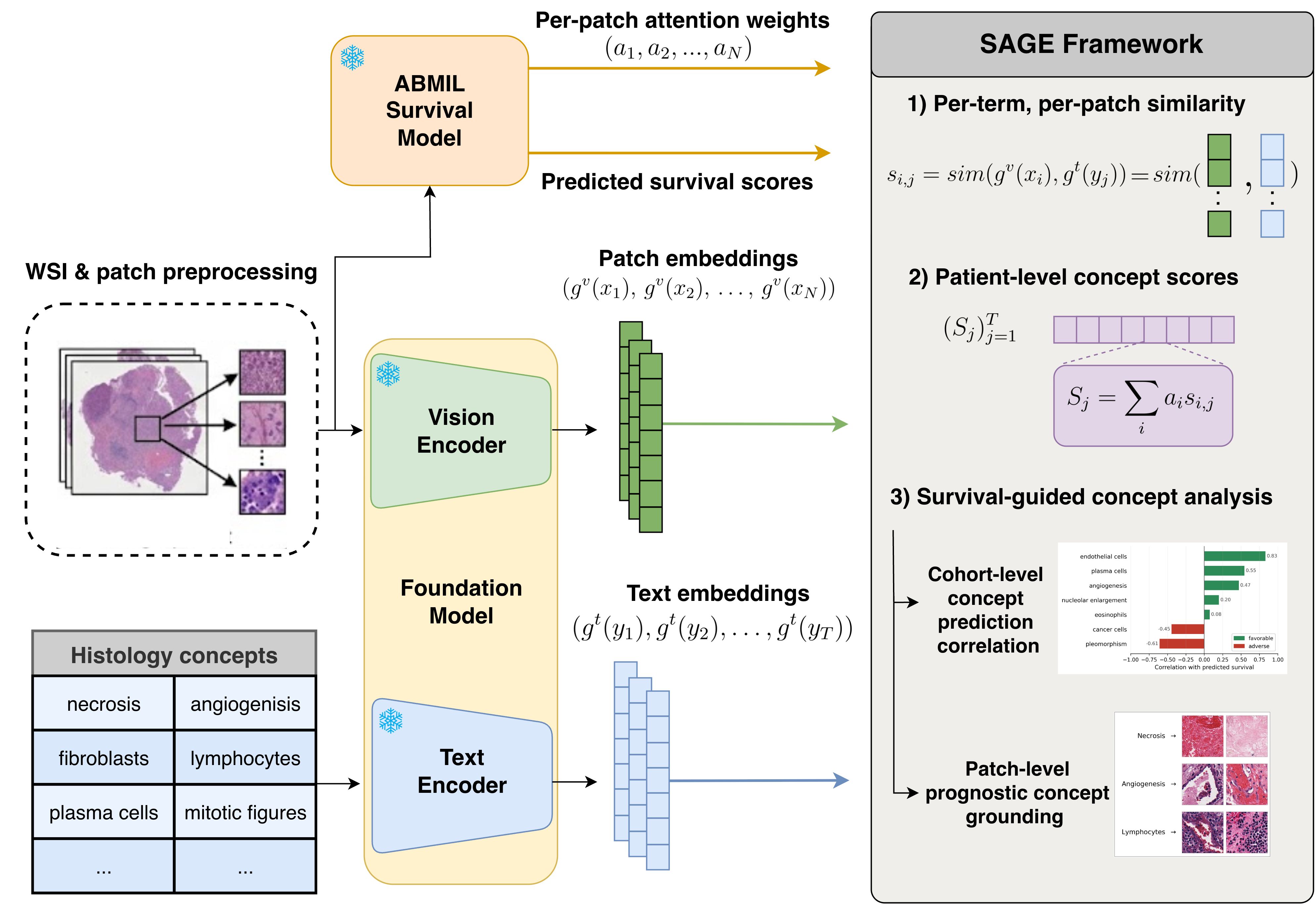}
\caption{SAGE: A framework for global semantic explanations of trained ABMIL models.}
\label{fig:pipeline}
\end{figure}


\subsubsection{Setup and notation.}
A WSI is represented as a bag of $N$ patches
$X = \{x_1,\dots,x_N\}$, which are processed by a frozen vision encoder (i.e. foundation model) $h^v$ to produce embeddings $h^v(x_i)$. An ABMIL network
assigns each patch an attention weight $a_i$ and aggregates patch-level representations into a
slide-level representation $z$:
\begin{equation}
a_i=\frac{\exp\!\big(w^\top\tanh(Vh^v(x_i))\big)}{\sum_{j=1}^N\exp\!\big(w^\top\tanh(Vh^v(x_j))\big)},
\qquad
z = \sum_{i=1}^N a_i h^v(x_i),
\end{equation}
with learnable parameters $w, V$. A downstream head then maps $z$ to a
slide-level prediction (e.g.\ a survival score). In SAGE we treat the entire
ABMIL predictor as frozen and use only its attention weights and output scores.

\subsubsection{Language-grounded semantic projection.}
We define a dictionary of $T=25$ pathology concepts that encompass common cell types
(e.g.\ \emph{lymphocytes}), cell descriptors (e.g.\ \emph{mitotic figures}), and
microenvironmental features (e.g.\ \emph{necrosis}) often used in interpreting
histology slides; the full list (\Cref{tab:terms}) was curated under pathologist
guidance. The framework is flexible to the terms included, and users can add or remove
terms as needed.

\begin{table}[t]
\caption{The $25$-term pathology concept dictionary, grouped by category.}
\label{tab:terms}
\centering
\setlength{\tabcolsep}{6pt}\small
\begin{tabular}{@{}>{\raggedright\arraybackslash}p{3.1cm} >{\raggedright\arraybackslash}p{8.3cm}@{}}
\toprule
Category & Terms \\
\midrule
Cell types & lymphocytes, plasma cells, macrophages, neutrophils, eosinophils,
fibroblasts, endothelial cells, epithelial cells, adipocytes, cancer cells \\[2pt]
Cell / nuclear descriptors & mitotic figures, apoptosis, pleomorphism, hyperchromasia,
nucleolar enlargement, dysplasia, poor differentiation \\[2pt]
Microenvironment \& architecture & angiogenesis, desmoplasia, scarring, necrosis,
lymphoid aggregates, hypercellularity, lymphovascular invasion, perineural invasion \\
\bottomrule
\end{tabular}
\end{table}
For each concept $y_j, j \in [1,T]$, the text encoder of a pathology VLM $g^t$ generates an embedding
$g^t(y_j)$. For robustness, we use a prompt-ensembling strategy, averaging
over $22$ natural-language templates provided by the CONCH repository \cite{lu2024conch}. For
each patch $x_i$, we use the VLM's vision encoder $g^v$ to compute the cosine similarity
$s_{i,j} = \langle g^v(x_i), g^t(y_j)\rangle/(\|g^v(x_i)\|\,\|g^t(y_j)\|)$ and aggregate it into an
attention-weighted, patient-level score $S_j = \sum_{i=1}^N a_i\,s_{i,j}$ using the attention weights of the ABMIL model.  
Intuitively, $S_j$ quantifies how strongly the ABMIL model attends to morphology
semantically related to concept $j$ on a slide, using a VLM to quantify patch-level semantic similarity. We note that the framing of our method allows SAGE to leverage different vision encoders for the ABMIL predictor and the vision-language cosine similarity computation. Our goal is to explain the ABMIL predictor itself, with SAGE transforming the localized attention weights into semantically meaningful scores.
\vspace{-5pt}
\subsubsection{From concept scores to global explanations.}
Given a trained ABMIL model and a concept dictionary, SAGE produces a vector $(S_j)_{j=1}^T$ of attention-weighted concept scores for each patient. To obtain
a global explanation of what the model has learned, we compare these scores
to the model’s own predictions across patients. In our experiments,
the ABMIL output is a continuous survival score, and we summarize the
association between each concept and model predictions using a rank
correlation coefficient over the cohort, as described below.
\subsubsection{Explaining prognostic models with SAGE.}
\vspace{-5pt}
For survival prediction, the slide representation $z$ is mapped to discrete-time
hazard logits $l_k$ over predefined time bins $k$:
\begin{equation}
h_k = \sigma(l_k), \qquad
P_k = \prod_{j \le k} (1 - h_j),
\end{equation}
where $h_k$ is the hazard and $P_k$ the discrete survival probability at bin $k$.
As is common in prognostication studies, we define a cumulative prognostic score
$\hat{\tau} = \sum_k P_k$, which increases with predicted survival time. The ABMIL
parameters are trained with a discrete-time negative log-likelihood survival
loss~\cite{zadeh2020nll}, and then frozen for the SAGE analysis.

\paragraph{Statistical association analysis.}
\label{subsec:stat}
For each concept $y_j$, we compute the Spearman rank correlation $r$ between $S_j$ and the ABMIL-predicted prognostic score $\hat{\tau}$ on test-set patients. We perform five-fold cross validation for all experiments, reporting the cross-fold mean
correlation with confidence intervals computed via bootstrapping and $p$-values computed via permutation ($5000$ resamples and permutations respectively, both performed within folds). Since $\hat{\tau}$ increases with predicted survival time, a
positive $r$ indicates that higher values of concept $y_j$ co-occur with
patients the model predicts to survive longer (favorable association), and a
negative $r$ indicates the opposite (adverse association).

\vspace{-5pt}
\section{Experiments and Results}

\subsection{Data and Setup}
We use seven TCGA cohorts (BRCA, BLCA, CESC, COAD, KIRC, LGG, LUAD), with
disease-specific survival (DSS) as the endpoint. We sample one slide per patient,
resulting in cohort sizes of $1014$, $370$, $246$, $398$, $479$, $478$ and $426$
respectively. Our core experiments use CONCH as both the vision
encoder for ABMIL fitting and explainability through SAGE. We additionally perform sensitivity analyses using
MUSK and UNI2. WSIs are processed and tiled using PathFMTools \cite{Diab2026-dk} with foreground tissue segmented via the HEST pipeline~\cite{jaume2024hest}.

\paragraph{ABMIL training.}
The discrete-time survival head uses four time bins set to the training-fold
event-time quartiles. Training is performed for up to $40$ epochs using the Adam optimizer \cite{Kingma2014-um} and early
stopping (patience $8$). Five-fold cross-validation is used, stratified by event-time quartile, where for each fold, three splits are used for training, one for validation, and one for testing. For each (cohort, model) pair, we sweep the same small grid:
learning rate $\in\{5\!\times\!10^{-5},10^{-4},2\!\times\!10^{-4}\}$ and hidden
dimension $\in\{256,384,512\}$. We then select the values leading to the highest
mean validation C-index across folds before proceeding to testing.

\subsection{Prognostic Performance}
An ABMIL predictor using the CONCH vision embeddings attains a mean test C-index of $0.625$ across the seven cohorts,
ranging from $0.77\pm0.03$ on KIRC to $0.53\pm0.09$ on BLCA. These
concordance values are in line with existing work evaluating foundation model-based ABMIL survival
models on TCGA~\cite{shazam2025}, providing a basis for our explainability analysis.

\subsection{Semantic Explainability}
\begin{figure}[t]
\centering
\includegraphics[width=\textwidth]{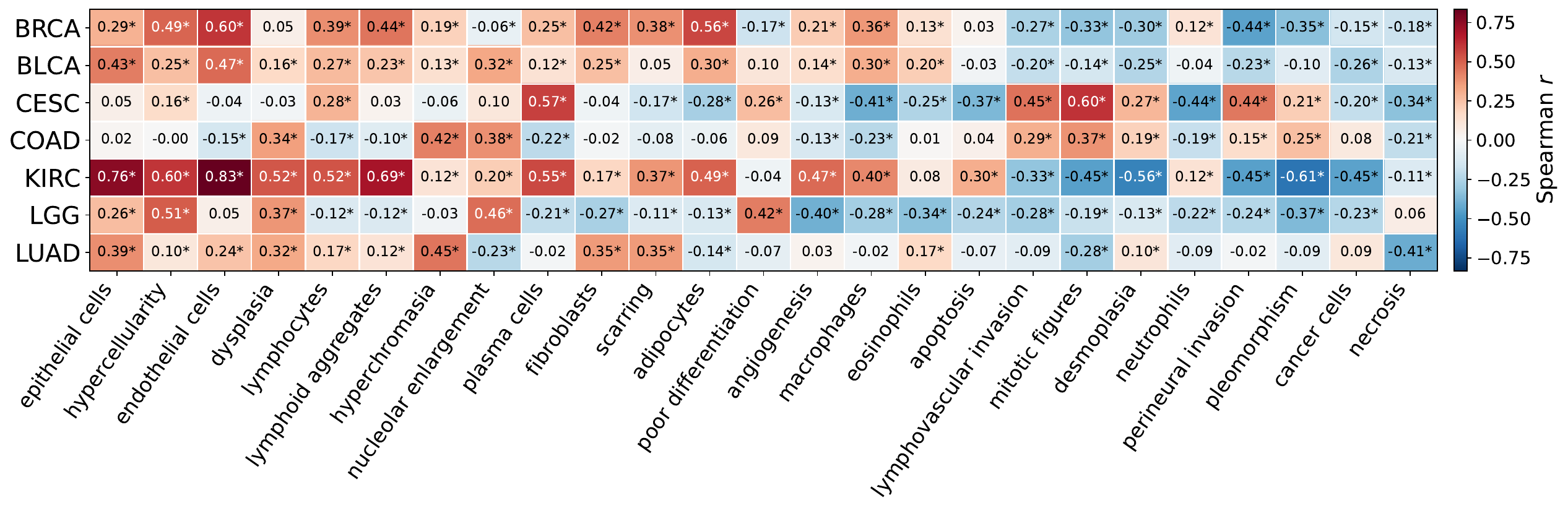}

\caption{SAGE profiles for prognosis prediction across seven TCGA cohorts using CONCH. Spearman $r$ between concept scores and predicted prognosis is shown per concept and cohort, with concepts sorted by mean $r$; positive values indicate correlation with longer predicted survival, negative values with shorter survival. Asterisks mark $p<0.05$ via permutation test.}
\label{fig:heatmap}
\end{figure}

The SAGE profile of the CONCH ABMIL predictor for each cohort is summarized in Fig.~\ref{fig:heatmap}, which shows the Spearman correlation between each concept and the prognosis predictions. Averaged across cohorts, the concept most associated with favorable survival predictions is \emph{epithelial cells}, with \emph{necrosis} most associated with adverse predictions. These findings are biologically intuitive as epithelial cells represent normal tissue, and necrosis indicates cell death, which has been shown to be negatively prognostic in several cancers \cite{Zhang2018-mv,Pollheimer2010-bd}. 

\begin{figure}[h]
\centering
\includegraphics[width=\textwidth]{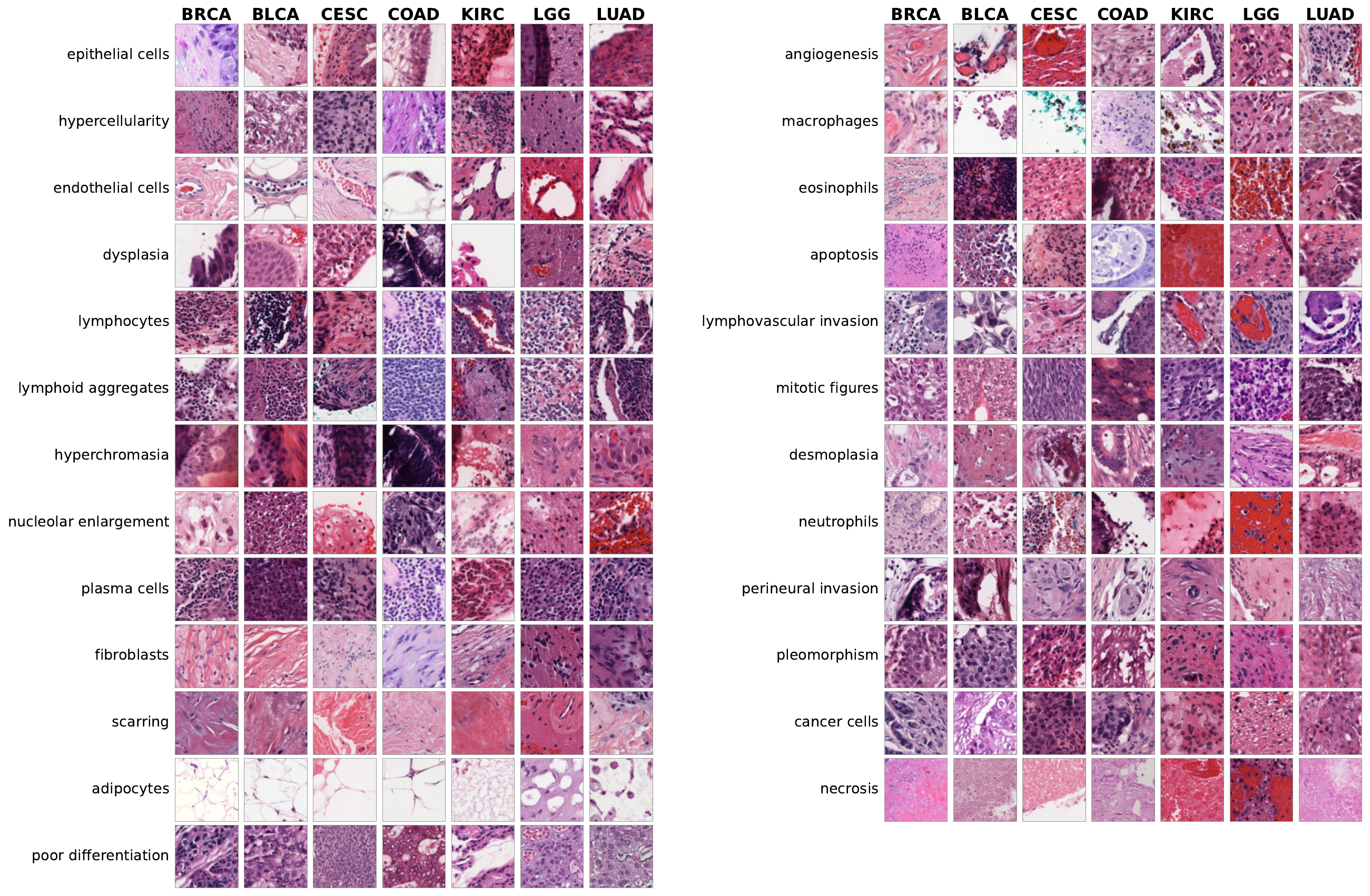}
\caption{Example patches for each concept. For each concept and cohort, a representative patch is sampled from slides with high concept scores and high ABMIL attention (top 10\% for each; based on CONCH).}
\label{fig:patches}
\end{figure}

Across cancer types, the most pronounced SAGE profile is observed in renal cell carcinoma (KIRC), which also corresponds to the highest prognostic performance.
Concept scores for \emph{endothelial cells} (cells that line blood vessels; $r=0.83\pm0.07$), \emph{plasma cells} (blood cells; $r=0.55\pm0.19$), and \emph{angiogenesis} (blood vessel growth; $r=0.47\pm0.11$) all correlate with longer predicted survival. This aligns with findings that angiogenic molecular subtypes of KIRC are associated with favorable prognosis~\cite{motzer2020rcc}. 
Another notable pattern is the strong negative association between \emph{neutrophils} and predicted prognosis in CESC (cervical squamous cell carcinoma and endocervical adenocarcinoma), also aligning with prior biomarker studies \cite{He2021-az}. 
Other associations are less clear-cut, such as that between \emph{hypercellularity} and longer predicted survival (second highest correlation across concepts). Hypercellularity can indicate high proliferation of cancer cells but may also reflect dense immune infiltration; inspection of example patches with high hypercellularity scores and high ABMIL attention revealed dense immune infiltrates (Fig.~\ref{fig:patches}), potentially explaining the favorable association.


\subsection{Dictionary Sufficiency: A Text-Only Survival Model}
\label{sec:dictablation}
A prerequisite for interpreting the concept scores as meaningful is that the $25$-term
dictionary itself captures prognostic signal. We tested this directly by training an
ABMIL survival model whose per-patch features are replaced by the patch-concept
similarities to the dictionary (one channel per prompt template of each term, in
CONCH's similarity space).
Despite this compression to an interpretable, concept-only representation,
the text-only model is competitive across all seven cohorts, achieving an average C-index that matches the vision-based model ($0.625$; Table \ref{tab:dict}). 
This performance thus validates the dictionary as a meaningful prognostic basis for explaining vision-based ABMIL models. The comparable prognostic accuracy further suggests that the text-based model could serve as a standalone alternative approach more broadly.

\begin{table}[h]
\caption{Prognostic performance on TCGA cohorts (test C-index, mean\,$\pm$\,std over 5 folds). The baseline ABMIL predictor based on CONCH vision embeddings aligns with performance reported in literature. Using the text-only similarity embeddings achieves similar performance. \textbf{Bold}
marks the better of the two per cohort.}
\label{tab:dict}
\centering
\setlength{\tabcolsep}{6pt}\small
\begin{tabular}{lccc}
\toprule
Cohort & $n$ & CONCH (vision) & text-only \\
\midrule
BRCA & 1014 & \textbf{0.612\,$\pm$\,0.06} & 0.582\,$\pm$\,0.07 \\
BLCA & 370  & 0.531\,$\pm$\,0.09 & \textbf{0.584\,$\pm$\,0.08} \\
CESC & 246  & 0.588\,$\pm$\,0.09 & \textbf{0.592\,$\pm$\,0.06} \\
COAD & 398  & 0.626\,$\pm$\,0.11 & \textbf{0.627\,$\pm$\,0.08} \\
KIRC & 479  & \textbf{0.774\,$\pm$\,0.03} & 0.760\,$\pm$\,0.03 \\
LGG  & 478  & \textbf{0.669\,$\pm$\,0.11} & 0.633\,$\pm$\,0.06 \\
LUAD & 426  & 0.577\,$\pm$\,0.06 & \textbf{0.597\,$\pm$\,0.05} \\
\midrule
Average &  & 0.625 & 0.625 \\
\bottomrule
\end{tabular}
\end{table}
\vspace{-3pt}
\subsection{Attention vs. Zero-Shot Aggregation: An Ablation}
\begin{figure}[t]
\centering
\includegraphics[width=\textwidth]{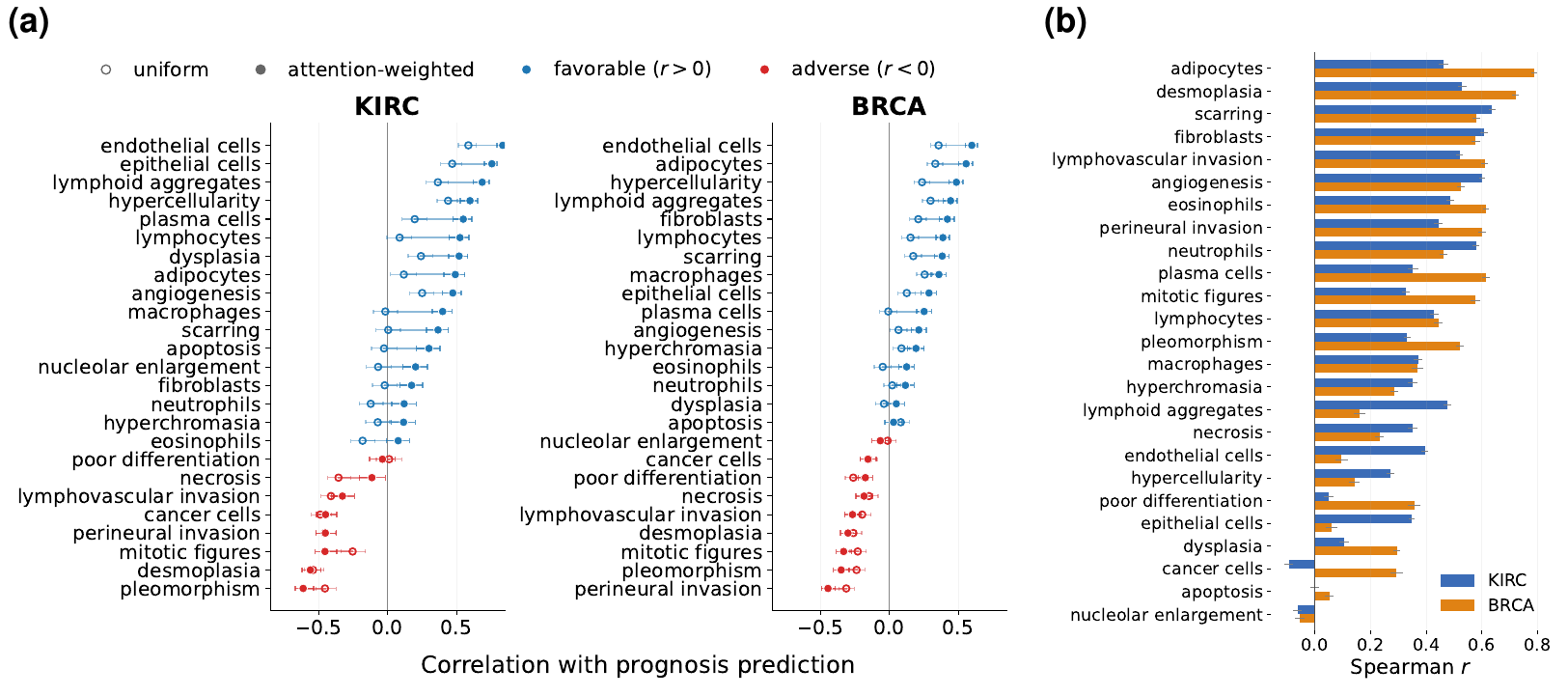}
\caption{\textbf{(a)} Attention ablation. Spearman $r$ between concept scores and prognosis prediction under uniform (open) vs.\ attention-weighted (filled) aggregation; 
bars = $95\%$ bootstrap CI. Attention sharpens the associations present in the uniform baseline. \textbf{(b)} Similarity in patch-level concept scores between CONCH and MUSK. For each concept, the mean per-slide Spearman
$r$ between CONCH and MUSK patch-level scores is displayed, separately for BRCA and KIRC. Architectural/stromal
concepts show higher concordance (top) than fine cytological ones (bottom).}
\label{fig:ablation}
\end{figure}

\vspace{-3pt}
Our semantic scores weight patch-concept similarities by the survival model's
attention, so a natural ablation asks whether that weighting adds anything over uniform (equal-weight) mean similarity.
We recomputed slide-level concept scores using uniform patch weights, and compared the resulting concept-predicted prognosis correlations.
We find that the correlative trends exist with uniform weighting, and become sharpened with attention weighting, as illustrated in Fig.~\ref{fig:ablation}a for KIRC and BRCA. 
Across all $175$
concept-cohort pairs attention-weighting increases $|r|$ for $75\%$ of them (mean
$\Delta|r|=+0.07$; one-sided paired Wilcoxon $p=4\times10^{-13}$). 
These results suggest that the attention-weighting better explains the model's predictions, as expected given ABMIL's formulation, and that the model may indeed be learning to focus on underlying discriminative features in the data.

\vspace{-5pt}
\subsection{Foundation Model Ablation: Does the Choice of VLM Matter?}
\label{sec:fmablation}
\vspace{-4pt}
We repeat the analysis with two further encoders, UNI2~\cite{chen2024uni} and MUSK~\cite{xiang2025musk}, on two cohorts (BRCA, the largest, and KIRC, the most prognostic). As UNI2 is a vision-only model, we use its features to train the ABMIL model, but score the attended patches against concepts in both CONCH's and MUSK's text-image spaces. The vision-based prognostic performance is similar between the three models (Table~\ref{tab:fm}). The text-only representation, however, is competitive only for CONCH: collapsing each patch to MUSK's $25$-concept similarities drops
concordance to $0.53$/$0.59$ on BRCA/KIRC. This result suggests that the predictive value of the interpretable space hinges on the quality of the VLM's text-image alignment, and supports the use of CONCH in our pipeline. Across all concepts, UNI2 tracks CONCH almost perfectly on KIRC ($r=0.98$ in rank order of concepts) and moderately on BRCA ($0.31$), suggesting that different ABMIL predictors may rely on both shared and distinct features. MUSK, with its own vision-language space, is more distinct from CONCH but still exhibits positive rank correlation ($r=0.48$/$0.19$ for KIRC/BRCA; see also
Fig.~\ref{fig:ablation}b). 

\begin{table}[t]
\caption{Foundation model ablation. Test C-index (mean\,$\pm$\,std over 5 folds) is displayed for BRCA and KIRC when using each model's vision encoder, and when using the text-only embedding variant for CONCH and MUSK. \textbf{Bold} = best per column.}
\label{tab:fm}
\centering
\setlength{\tabcolsep}{6pt}\small
\begin{tabular}{lcccc}
\toprule
& \multicolumn{2}{c}{BRCA} & \multicolumn{2}{c}{KIRC} \\
\cmidrule(lr){2-3}\cmidrule(lr){4-5}
Encoder & vision & text-only & vision & text-only \\
\midrule
CONCH & 0.612\,$\pm$\,0.06 & \textbf{0.582\,$\pm$\,0.07} & \textbf{0.774\,$\pm$\,0.03} & \textbf{0.760\,$\pm$\,0.03} \\
UNI2  & \textbf{0.659\,$\pm$\,0.07} & --- & 0.751\,$\pm$\,0.05 & --- \\
MUSK  & 0.605\,$\pm$\,0.11 & 0.527\,$\pm$\,0.09 & 0.740\,$\pm$\,0.03 & 0.595\,$\pm$\,0.15 \\
\bottomrule
\end{tabular}
\end{table}

\vspace{-5pt}
\section{Discussion}

\subsubsection{Limitations and future work.} Concept associations are measured against the ABMIL model's predictions, so they should be interpreted as hypothesis-generating rather than causal. While the use of a VLM provides a scalable means to assess concept similarity, the model may make errors and have non-uniform accuracy across concepts. The default list of 25 concepts may miss cancer-specific or subtle morphological features, so users are encouraged to customize this list for specific applications. We tested seven TCGA cohorts, a common prognostication benchmark, but future validation in other cohorts is important. 
\vspace{-5pt}
\subsubsection{Conclusion.} SAGE is a lightweight, plug-and-play framework that translates ABMIL
attention into global concept associations. Applied across seven TCGA cohorts and
three foundation models, the recovered profiles often align with biological knowledge and are strongest where the model is most accurate (notably KIRC). Through semantic explainability, SAGE may help foster physician trust in emerging prognostic applications while also advancing our understanding of the biological relationships between histological features and patient outcomes.
\vspace{-5pt}

\begin{credits}
\subsubsection{Code Availability.}Code is available at \url{https://github.com/lotterlab/sage}.

\subsubsection{\ackname}W.L. acknowledges funding support from the Ellison Foundation, the Wong Family Award, the Louis B. Mayer Foundation, the National Institute of Biomedical Imaging and Bioengineering award R21EB035247, and the National Library of Medicine award R01LM014775.

\subsubsection{\discintname}
The authors have no competing interests to declare.
\end{credits}

\bibliographystyle{splncs04}
\bibliography{references}

@ARTICLE{Koh2020-ti,
  title    = "Concept Bottleneck Models",
  author   = "Koh, Pang Wei and Nguyen, Thao and Tang, Yew Siang and Mussmann,
              Stephen and Pierson, Emma and Kim, Been and Liang, Percy",
  journal  = "ICML",
  year     =  2020
}

@ARTICLE{shazam2025,
  title         = "Unifying multiple foundation models for advanced
                   computational pathology",
  author        = "Lei, Wenhui and Tan, Yusheng and Li, Anqi and Chen, Hanyu and
                   Tian, Hengrui and Li, Ruiying and Jiang, Zhengqun and Yan,
                   Fang and Zhang, Xiaofan and Zhang, Shaoting",
  journal       = "arXiv",
  month         =  mar,
  year          =  2025,
  archivePrefix = "arXiv",
  primaryClass  = "cs.CV"
}

@ARTICLE{nguyen2025vleer,
  title         = "{VLEER}: Vision and Language Embeddings for Explainable whole
                   slide image representation",
  author        = "Nguyen, Anh Tien and Byeon, Keunho and Kim, Kyungeun and
                   Kwak, Jin Tae",
  journal       = "arXiv",
  month         =  feb,
  year          =  2025,
  archivePrefix = "arXiv",
  primaryClass  = "cs.CV"
}

@ARTICLE{sun2025conceptmil,
  title         = "Label-free concept based multiple Instance Learning for
                   gigapixel histopathology",
  author        = "Sun, Susu and Tessier, Leslie and Meeuwsen, Frédérique and
                   Grisi, Clément and van Midden, Dominique and Litjens, Geert
                   and Baumgartner, Christian F",
  journal       = "arXiv [cs.CV]",
  month         =  jan,
  year          =  2025,
  archivePrefix = "arXiv",
  primaryClass  = "cs.CV"
}

@ARTICLE{zhao2024conceppath,
  title     = "Aligning knowledge concepts to whole slide images for precise
               histopathology image analysis",
  author    = "Zhao, Weiqin and Guo, Ziyu and Fan, Yinshuang and Jiang, Yuming
               and Yeung, Maximus C F and Yu, Lequan",
  journal   = "NPJ Digit. Med.",
  publisher = "Springer Science and Business Media LLC",
  volume    =  7,
  number    =  1,
  pages     =  383,
  month     =  dec,
  year      =  2024,
  language  = "en"
}

@ARTICLE{liu2025vlsa,
  title   = "Interpretable Vision-Language Survival Analysis with Ordinal
             Inductive Bias for Computational Pathology",
  author  = "Liu, Pei and Ji, Luping and Gou, Jiaxiang and Fu, Bo and Ye, Mao",
  journal = "International Conference on Learning Representations",
  volume  =  2025,
  pages   = "99062--99090",
  month   =  may,
  year    =  2025
}

@ARTICLE{motzer2020rcc,
  title     = "Molecular subsets in renal cancer determine outcome to checkpoint
               and angiogenesis blockade",
  author    = "Motzer, Robert J and Banchereau, Romain and Hamidi, Habib and
               Powles, Thomas and McDermott, David and Atkins, Michael B and
               Escudier, Bernard and Liu, Li-Fen and Leng, Ning and Abbas,
               Alexander R and Fan, Jinzhen and Koeppen, Hartmut and Lin,
               Jennifer and Carroll, Susheela and Hashimoto, Kenji and
               Mariathasan, Sanjeev and Green, Marjorie and Tayama, Darren and
               Hegde, Priti S and Schiff, Christina and Huseni, Mahrukh A and
               Rini, Brian",
  journal   = "Cancer Cell",
  publisher = "Elsevier BV",
  month     =  dec,
  year      =  2020,
  language  = "en"
}

@ARTICLE{jaume2024hest,
  title         = "{HEST}-{1k}: A Dataset for Spatial Transcriptomics and
                   Histology Image Analysis",
  author        = "Jaume, Guillaume and Doucet, Paul and Song, Andrew H and Lu,
                   Ming Y and Almagro-Pérez, Cristina and Wagner, Sophia J and
                   Vaidya, Anurag J and Chen, Richard J and Williamson, Drew F K
                   and Kim, Ahrong and Mahmood, Faisal",
  journal       = "arXiv",
  year          =  2024,
  archivePrefix = "arXiv",
  primaryClass  = "cs.CV",
  language      = "en"
}

@ARTICLE{zadeh2020nll,
  title     = "Bias in cross-entropy-based training of deep survival networks",
  author    = "Zadeh, Shekoufeh Gorgi and Schmid, Matthias",
  journal   = "IEEE Trans. Pattern Anal. Mach. Intell.",
  publisher = "Institute of Electrical and Electronics Engineers (IEEE)",
  volume    =  43,
  number    =  9,
  pages     = "3126--3137",
  month     =  sep,
  year      =  2021,
  language  = "en"
}

@INPROCEEDINGS{ilse2018attention,
  title     = "Attention-based Deep Multiple Instance Learning",
  author    = "Ilse, Maximilian and Tomczak, Jakub and Welling, Max",
  booktitle = "International Conference on Machine Learning",
  publisher = "PMLR",
  month     =  jul,
  year      =  2018,
  language  = "en"
}

@ARTICLE{He2021-az,
  title     = "The prognostic significance of tumor-infiltrating lymphocytes in
               cervical cancer",
  author    = "He, Mengdi and Wang, Yiying and Zhang, Guodong and Cao, Kankan
               and Yang, Moran and Liu, Haiou",
  journal   = "J. Gynecol. Oncol.",
  publisher = "Asian Society of Gynecologic Oncology; Korean Society of
               Gynecologic Oncology and Colposcopy",
  volume    =  32,
  number    =  3,
  pages     = "e32",
  month     =  may,
  year      =  2021,
  language  = "en"
}

@ARTICLE{Pollheimer2010-bd,
  title     = "Tumor necrosis is a new promising prognostic factor in
               colorectal cancer",
  author    = "Pollheimer, Marion J and Kornprat, Peter and Lindtner, Richard A
               and Harbaum, Lars and Schlemmer, Andrea and Rehak, Peter and
               Langner, Cord",
  journal   = "Hum. Pathol.",
  publisher = "Elsevier BV",
  volume    =  41,
  number    =  12,
  pages     = "1749--1757",
  month     =  dec,
  year      =  2010,
  language  = "en"
}

@ARTICLE{Zhang2018-mv,
  title     = "Tumor necrosis as a prognostic variable for the clinical outcome
               in patients with renal cell carcinoma: a systematic review and
               meta-analysis",
  author    = "Zhang, Lijin and Zha, Zhenlei and Qu, Wei and Zhao, Hu and Yuan,
               Jun and Feng, Yejun and Wu, Bin",
  journal   = "BMC Cancer",
  publisher = "Springer Science and Business Media LLC",
  volume    =  18,
  number    =  1,
  pages     = "870",
  month     =  sep,
  year      =  2018,
  language  = "en"
}

@ARTICLE{chen2024uni,
  title     = "Towards a general-purpose foundation model for computational
               pathology",
  author    = "Chen, Richard J and Ding, Tong and Lu, Ming Y and Williamson,
               Drew F K and Jaume, Guillaume and Song, Andrew H and Chen, Bowen
               and Zhang, Andrew and Shao, Daniel and Shaban, Muhammad and
               Williams, Mane and Oldenburg, Lukas and Weishaupt, Luca L and
               Wang, Judy J and Vaidya, Anurag and Le, Long Phi and Gerber,
               Georg and Sahai, Sharifa and Williams, Walt and Mahmood, Faisal",
  journal   = "Nat. Med.",
  publisher = "Nature Publishing Group",
  volume    =  30,
  number    =  3,
  pages     = "850--862",
  month     =  mar,
  year      =  2024,
  language  = "en"
}

@ARTICLE{xiang2025musk,
  title     = "A vision-language foundation model for precision oncology",
  author    = "Xiang, Jinxi and Wang, Xiyue and Zhang, Xiaoming and Xi, Yinghua
               and Eweje, Feyisope and Chen, Yijiang and Li, Yuchen and
               Bergstrom, Colin and Gopaulchan, Matthew and Kim, Ted and Yu,
               Kun-Hsing and Willens, Sierra and Olguin, Francesca Maria and
               Nirschl, Jeffrey J and Neal, Joel and Diehn, Maximilian and Yang,
               Sen and Li, Ruijiang",
  journal   = "Nature",
  publisher = "Springer Science and Business Media LLC",
  volume    =  638,
  number    =  8051,
  pages     = "769--778",
  month     =  feb,
  year      =  2025,
  language  = "en"
}

@ARTICLE{lu2024conch,
  title     = "A visual-language foundation model for computational pathology",
  author    = "Lu, Ming Y and Chen, Bowen and Williamson, Drew F K and Chen,
               Richard J and Liang, Ivy and Ding, Tong and Jaume, Guillaume and
               Odintsov, Igor and Le, Long Phi and Gerber, Georg and Parwani,
               Anil V and Zhang, Andrew and Mahmood, Faisal",
  journal   = "Nat. Med.",
  publisher = "Nature Publishing Group",
  year      =  2024,
  language  = "en"
}

@INPROCEEDINGS{Diab2026-dk,
  title     = "Leveraging Foundation Models for Histological Grading in
               Cutaneous Squamous Cell Carcinoma using {PathFMTools}",
  author    = "Diab, Abdul Rahman and Karn, Emily E and Wu, Renchin and Ruiz,
               Emily S and Lotter, William",
  booktitle = "Machine Learning for Health Symposium",
  publisher = "PMLR",
  month     =  may,
  year      =  2026,
  language  = "en"
}

@ARTICLE{Kingma2014-um,
  title         = "Adam: A method for stochastic optimization",
  author        = "Kingma, Diederik P and Ba, Jimmy",
  journal       = "arXiv",
  year          =  2014,
  archivePrefix = "arXiv",
  primaryClass  = "cs.LG"
}

\end{document}